\documentclass[runningheads]{llncs}
\usepackage[latin9]{inputenc}
\usepackage{amsmath}
\usepackage{graphicx}

\makeatletter

\providecommand{\tabularnewline}{\\}

\makeatother

\begin{document}
\title{Weakly supervised one-stage vision and language disease detection
using large scale pneumonia and pneumothorax studies }
\titlerunning{LITERATI: weakly-supervised one-stage detection}
\author{Leo K. Tam\inst{1} \and Xiaosong Wang\inst{1} \and Evrim Turkbey\inst{2}\and
Kevin Lu\inst{1}\and Yuhong Wen\inst{1} \and Daguang Xu\inst{1} }
\authorrunning{Tam, L.K., Wang, X, et. al.}
\institute{NVIDIA, 2788 San Tomas Expy, Santa Clara, CA, 95051 \and National
Institute of Health Clinical Center, 10 Center Dr., Bethesda, MD 20814}
\maketitle
\begin{abstract}
Detecting clinically relevant objects in medical images is a challenge
despite large datasets due to the lack of detailed labels. To address
the label issue, we utilize the scene-level labels with a detection
architecture that incorporates natural language information. We present
a challenging new set of radiologist paired bounding box and natural
language annotations on the publicly available MIMIC-CXR dataset especially
focussed on pneumonia and pneumothorax. Along with the dataset, we
present a joint vision language weakly supervised transformer layer-selected
one-stage dual head detection architecture (LITERATI) alongside strong
baseline comparisons with class activation mapping (CAM), gradient
CAM, and relevant implementations on the NIH ChestXray-14 and MIMIC-CXR
dataset. Borrowing from advances in vision language architectures,
the LITERATI method demonstrates joint image and referring expression
(objects localized in the image using natural language) input for
detection that scales in a purely weakly supervised fashion. The architectural
modifications address three obstacles -- implementing a supervised
vision and language detection method in a weakly supervised fashion,
incorporating clinical referring expression natural language information,
and generating high fidelity detections with map probabilities. Nevertheless,
the challenging clinical nature of the radiologist annotations including
subtle references, multi-instance specifications, and relatively verbose
underlying medical reports, ensures the vision language detection
task at scale remains stimulating for future investigation. 

\end{abstract}
\section{Introduction}

Recently, the release of large scale datasets concerning chest x-rays
has enabled methods that scale with such datasets \cite{mimic-cxr,chexnet,chestx8}.
Whereas image classification implementations may reach adequate performance
using scene-level labels, significantly more effort is required for
annotation of bounding boxes around numerous visual features of interest.
Yet there is detailed information present in released clinical reports
that could inform natural language (NL) methods. The proposed method
brings together advances in object detection \cite{speedacctrade},
language \cite{bertart}, and their usage together \cite{tienet,onestage}. 

Typically object detection algorithms are either multi-stage with
a region proposal stage or single stage with proposed regions scored
to a certain object class when proposed \cite{yolo,frcnn}. The single
stage detectors have the benefit of fast inference time at often nearly
the same performance in accuracy \cite{yolo,yolov3}. The object detection
networks benefit from using the same classification network architecture
as their image classification cousins, where the visual features carry
significance and shared modularity of networks holds. The modularity
of network architectures is further realized with recent vision language
architectures.

Vision language networks seek to encode the symbolic and information
dense content in NL with visual features to solve applications such
as visual question and answering, high fidelity image captioning,
and other multi-modal tasks, some of which have seen application in
medical imaging \cite{tienet,deeplesion,wentao,ibm-morardi}. Recent
advances in NLP incorporate the transformer unit architecture, a computational
building block that allows the attention of every word to learn an
attentional weighting with regards to every other word in a sequence,
given standard NLP tasks such as cloze, next sentence prediction,
etc \cite{attentionall}. Furthermore, deep transformer networks of
a dozen or more layers trained for the language modeling task (next
word prediction) were found to be adaptable to a variety of tasks,
in part due to their ability to learn the components of a traditional
NLP processing pipeline \cite{bertnlppipe}. The combination of NLP
for the vision task of object detection centers around the issue of
visual grounding, namely given a referring phrase, how the phrase
places an object in the image. The computational generation of referring
phrases in a natural fashion is a nontrivial problem centered on photographs
of cluttered scenes, where prevailing methods are based on probabilistically
mapped potentials for attribute categories \cite{referit}. Related
detection methods on cluttered scenes emphasize a single stage and
end-to-end training \cite{referit,cvprdense}. 

In particular, our method builds on a supervised single stage visual
grounding method that incorporates fixed transformer embeddings \cite{onestageart}.
The original one-stage methods fuses the referring expression in the
form of a transformer embedding to augment the spatial features in
the YOLOv3 detector. Our method alters to a weakly supervised implementation,
taking care to ensure adequate training signal can be propagated to
the object detection backbone through the technique adapted for directly
allowing training signal through a fixed global average pooling layer
\cite{networkinnet}. The fast and memory efficient backbone of the
DarkNet-53 architecture combines with fixed bidirectionally encoded
features \cite{bertart} to visually ground radiologist-annotated
phrases. The fixed transformer embeddings are allowed increased flexibility
through a learned selection layer as corroborated by the concurrent
work \cite{alarm}, though here our explicit reasoning is to boost
the NL information (verified by ablation study) of a more sophisticated
phrasing then encountered in the generic visual grounding setting.
To narrow our focus for object detection, we consider two datasets
-- the ChestXray-14 dataset \cite{chestx8}, which was released with
984 bounding box annotations spread across 8 labels, and the MIMIC-CXR
dataset \cite{mimic-cxr}, for which we collected over 400 board-certified
radiologist bounding box annotations. Our weakly supervised transformer
layer-selected one-stage dual head detection architecture (LITERATI)
forms a strong baseline on a challenging set of annotations with scant
information provided by the disease label.

\section{Methods}

\begin{figure}
\centering{}\includegraphics[scale=0.32]{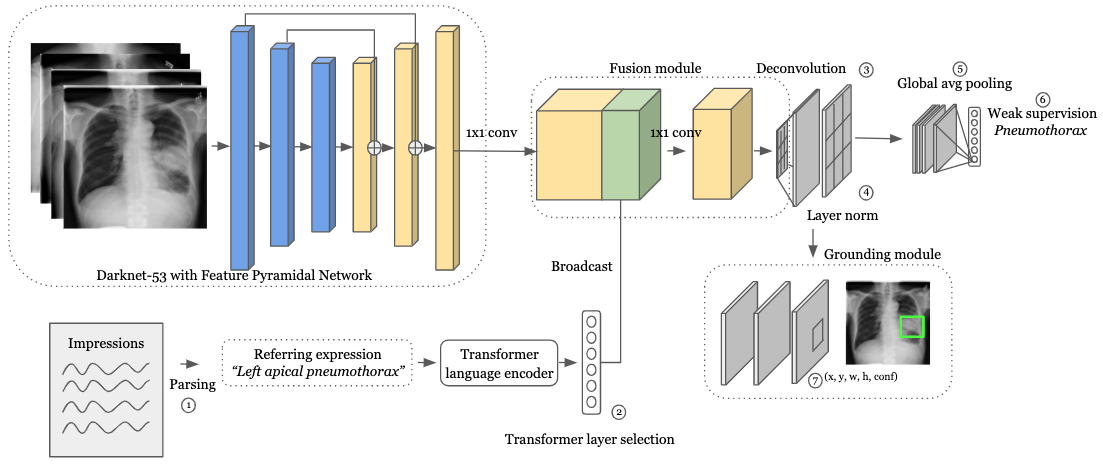}\caption{The LITERATI network architecture is a vision language one-stage detector
adapted from the supervised method of \cite{onestage} for the weakly
supervised case of medical imaging visually grounded object detection.
The changes are highlighted by numbers and described in Sec. \ref{subsec:Network-architecture}.}
\label{fig1}
\end{figure}

The architecture of our method is presented in Fig. \ref{fig1} with
salient improvements numbered. The inputs of the model are anteroposterior
view chest x-ray images and a referring expression parsed from the
associated clinical report for the study. The output of the model
is a probability map for each disease class (pneumonia and pneumothorax)
as well as a classification for the image to generate the scene level
loss. Intersection over union (IOU) is calculated for the input and
ground truth annotation as $\frac{A\cap B}{A\cup B}$ where A is the
input and B is the ground truth annotation. 

\subsection{Preprocessing}

The MIMIC-CXR dataset second stage release \cite{mimic-cxr} included
a reference label file built on the CheXpert labeler \cite{chexpert},
which we used for our filtering and data selection. The CheXpert labels
are used to isolate pneumonia and pneumothorax images and the corresponding
chest x-ray reports retrieved by subject and study number. The images
are converted from the full resolution (typically 2544x3056) to 416x416
to match the preferred one-stage resolution. For the ChestXray-14
dataset, the 1024x1024 PNG files are converted in PNG format.

For the MIMIC-CXR dataset, the radiologist reports are parsed to search
for referring expressions, i.e. an object of interest is identified
and located in the image \cite{referit}. The referring expression
is created using tools adapted from \cite{referit}. Namely, the tooling
uses the Stanford CoreNLP \cite{corenlp} parser and the NLTK tokenizer
\cite{nltk} to separate sentences into the R1-R7 attributes and reformed
where there is an object in the image as either subject or direct
object. Specifically, the referring phrase consists of the R7 attributes
(generics), R1 attributes (entry-level name), R5 attributes (relative
location), and finally R6 attributes (relative object) \cite{referit}.
Sample referring phrases in the reports are ``confluent opacity at
bases'', ``left apical pneumothorax'', and ``multifocal bilateral
airspace consolidation''. As occasionally, the referring phrase does
not capture the disease focus, the reports are additionally processed
to excerpt phrases with ``pneumonia'' and ``pneumothorax'' to
create a disease emphasis dataset split. For example, a phrase that
does not qualify as a canonical referring expression but is present
for the disease is, ``vague right mid lung opacity, which is of uncertain
etiology, although could represent an early pneumonia'' which is
a positive mention, standing in contrast to, ``no complications,
no pneumothorax'' as a negative mention. To include the presence
of normal and negative example images, data negative for pneumothorax
and pneumonia was mixed in at an equal ratio to positive data for
either category. The experiments on the disease emphasis phrases are
presented as an interrogative data distribution ablation of the NL
function. To further capture the language function, the scene level
label can be submitted for the language embedding itself. For the
ChestXray-14 dataset, the scene level label is the only textual input
publicly available.

During the NL ablation, experiments are performed on the MIMIC-CXR
dataset with three different levels of referring phrases provided
during training. The tersest level of phrasing is simply the scene
level label, which include the cases pneumonia, pneumothorax, pneumonia
and pneumothorax, or the negative phrase 'no pneumo'. The second level
and third level are the phrasing described previously. At test time,
the full annotation provided by the radiologist is used. 

Once the referring expressions are in place, the ingredients are set
for board-certified radiologist clinical annotations. We pair the
images and highlight relevant phrases in the clinical report by building
functionality on the publicly available MS COCO annotator \cite{mscocoanno}.
The radiologist is given free rein to select phrases that are clinically
relevant in the report to highlight. The radiologist has annotated
455 clinically relevant phrases with bounding boxes on the MIMIC-CXR
dataset, which we release at https://github.com/leotam/MIMIC-CXR-annotations.
As of writing, the annotations constitute the largest disease focussed
bounding box labels with referring expressions publicly released,
and we hope is a valuable contribution to the clinical visual grounding
milieu.

\subsection{Network architecture\label{subsec:Network-architecture}}

There are six modifications to \cite{onestage} noted in Fig. \ref{fig1}
beyond the parsing discussed. To adapt the network from supervised
to weakly supervised, the classification layer must be not be trained
to reduce partitioning of the data information. To achieve that purpose,
a global average pooling layer \cite{networkinnet} was combined with
a cross-entropy loss on the scene labels, Fig. \ref{fig1} (5, 6),
to replace the convolution-batch norm-convolution (CBC) layer that
generates the bounding box anchors in the original implementation.
To replace the CBC layer, a deconvolution layer norm layer was prepended
to the pooling layer Fig. \ref{fig1} (3), which additionally allowed
grounding on an image scale probability map Fig. \ref{fig1} (7),
instead of the anchor shifts typically dictated by the YOLOv3 implementation
\cite{yolov3}. 

For the NL implementation, the ability of transformer architectures
to implement aspects of the NLP pipeline \cite{bertnlppipe} suggests
a trained layer may be successful in increasing the expressivity of
the language information for the task. Normally, a fully connected
layer is appended to a transformer model to fine-tune for a given
task for various classification tasks \cite{squad2}. The architecture
includes a convolutional 1D layer in the language mapping module Fig.
\ref{fig1} (2) that allows access to all the transformer layers instead
of the linear layers on the averaged last four transformer layers
output in the original implementation \cite{onestage}. Such a modification
echos the custom attention on layers mechanism from a concurrent study
on the automatic labeler work \cite{alarm}. 

For the bounding box generation, we move to the threshold detection
method similar to \cite{chestx8}, which differs from the threshold
detection method in \cite{onestage}. The current generation method
uses the tractable probability map output after the deconvolution-layer
norm stage close to the weak supervision signal to select regions
of high confidence given a neighborhood size. Specifically, a maximal
filter is applied to the map probabilities as follows:
\begin{equation}
M=\left\{ \frac{e^{c_{i}}}{\sum_{i}e^{c_{i}}}\mid c_{i}\in C\right\} 
\end{equation}
\begin{equation}
S=\left\{ m\mid\max(m\in M)\,\forall\,m\mid\left\Vert m-x_{0}\right\Vert <d\right\} 
\end{equation}
\begin{equation}
x_{0}\equiv\frac{\overset{N}{\sum}(x-x_{0})}{N}=d\mid x\in S.
\end{equation}

First the M map probabilities are generated from the convolutional
outputs C via softmax, followed by maximal filtering (with the threshold
d as a hyperparameter set by tree-structured parzen estimator \cite{DBLP:conf/nips/BergstraBBK11},
as are all hyperparameters across methods) to generate the regions
S, and then $x_{0}$ center of masses collected as bounding box centers.
The original method \cite{yolov3} used the confidence scores to assign
a probability to a superimposed anchor grid. As the LITERATI output
is at parity resolution with the image, the deconvolution and maximal
filtering obviates the anchor grid and offset prediction mechanism.

\subsection{Training}

The implementation in PyTorch \cite{pytorchart} allows for straight-forward
data parallelism via the nn.DataParallel class, which we found was
key to performance in a timely fashion. The LITERATI network was trained
for 100 epochs (as per the original implementation) on a NVIDIA DGX-2
(16x 32 GB V100) with 16-way data parallelism using a batch size of
416 for approximately 12 hours. The dataset for the weakly supervised
case was split in an 80/10/10 ratio for training, test, and validation
respectively, or 44627, 5577, and 5777 images respectively. The test
set was checked infrequently for divergence from the validation set.
The epoch with the best validation performance was selected for the
visual grounding task and was observed to always perform better than
the overfitted last epoch. For the supervised implementation of the
original, there were not enough annotations and instead the default
parameters from \cite{onestage} were used.

\section{Results and discussion}

In Tab. \ref{tab1}, experiments are compared across the ChestXray-14
dataset. The methods show the gradient CAM implementation derived
from \cite{gradcam,thtang} outperforms the traditional CAM implementation
\cite{cam}, likely due to the higher resolution localization map.
Meanwhile the LITERATI with scene label method improves on the grad
CAM method most significantly at IOU = 0.2, though lagging at IOU=0.4
and higher. Above both methods is the jointly trained supervised and
unsupervised method from \cite{thoracicart}. The LITERATI method
and the one-stage method were designed for visual grounding, but the
performance for detection on the supervised one-stage method with
minimal modification is not strong at the dataset size here, 215 ChestXray-14
annotations for specifically pneumonia and pneumothorax as that is
the supervision available.
\begin{table}
\begin{centering}
\caption{ChestXray-14 detection accuracy}
\label{tab1} %
\begin{tabular}{|c|c|c|c|c|c|}
\hline 
IOU & 0.1 & 0.2 & 0.3 & 0.4 & 0.5\tabularnewline
\hline 
\hline 
CAM \cite{cam} WS & 0.505 & 0.290 & 0.150 & 0.075 & 0.030\tabularnewline
\hline 
Multi-stage S + WS \cite{thoracicart} & \textbf{0.615} & \textbf{0.505} & \textbf{0.415} & \textbf{0.275} & \textbf{0.180}\tabularnewline
\hline 
\hline 
Gradient CAM WS & 0.565 & 0.298 & 0.175 & \textbf{0.097} & \textbf{0.049}\tabularnewline
\hline 
LITERATI SWS & \textbf{0.593} & \textbf{0.417} & \textbf{0.204} & 0.088 & 0.046\tabularnewline
\hline 
\hline 
One-stage S & 0.115 & 0.083 & 0.073 & 0.021 & 0.003\tabularnewline
\hline 
\end{tabular}
\par\end{centering}
\centering{}Supervised (S), weakly supervised (WS), and scene-level
NLP (SWS) methods
\end{table}

\begin{figure}
\includegraphics[scale=0.29]{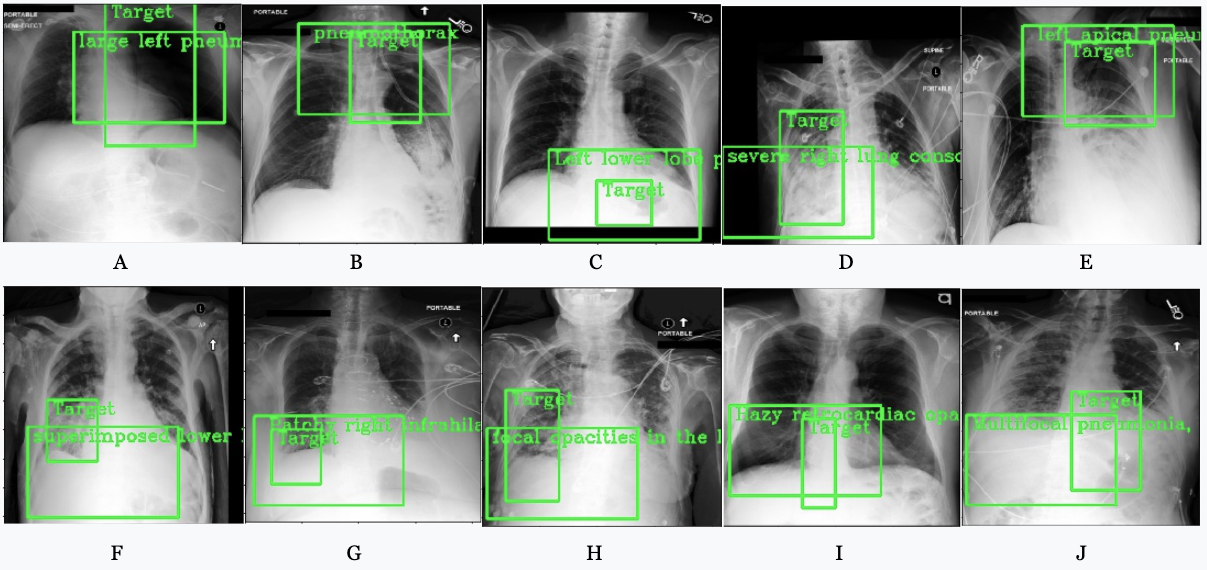}\caption{Qualitative detection results using the LITERATI method. The referring
expressions (A-J) are ``large left pneumothorax'', ``pneumothorax'',
``left lower lobe pneumonia'', ``severe right lung consolidation'',
``left apical pneumothorax'', ``superimposed lower lobe pneumonia'',
``patchy right infrahilar opacity'', ``focal opacities in the lateral
right mid lung and medial right lower lung'', ``hazy retrocardial
opacity'', and ``multifocal pneumonia''. In the top row (A-E),
mainly disease focussed referring phrases are localized. In the bottom
row (F-J), examples of difficult annotations where the referenced
location may be vague or spatially ambiguous are presented. }
\label{fig2}
\end{figure}

While we include two relevant prior art baselines without language
input, the present visual grounding task is more specifically contained
by the referring expression label and at times further removed from
the scene level label due to an annotation's clinical importance determined
by the radiologist. The data in Tab. \ref{tab2} shows detection accuracy
on the visual grounding task. The LITERATI method improves on the
supervised method at IOU=0.1 and on gradient CAM at all IOUs. 
\begin{table}
\begin{centering}
\caption{MIMIC-CXR detection accuracy}
\label{tab2} %
\begin{tabular}{|c|c|c|c|c|c|}
\hline 
IOU & 0.1 & 0.2 & 0.3 & 0.4 & 0.5\tabularnewline
\hline 
\hline 
Gradient CAM WS & 0.316 & 0.104 & 0.049 & 0.005 & 0.001\tabularnewline
\hline 
LITERATI NWS & 0.349 & 0.125 & 0.060 & 0.024 & 0.007\tabularnewline
\hline 
\hline 
One-stage S & 0.209 & 0.125 & 0.125 & 0.125 & 0.031\tabularnewline
\hline 
\end{tabular}
\par\end{centering}
\centering{}Supervised (S), weakly supervised (WS), and NLP (NWS)
methods
\end{table}
 We present qualitative results in fig. \ref{fig2} that cover a range
of annotations from disease focussed (``large left pneumonia'')
to more subtle features (``patchy right infrahilar opacity''). Of
interest are the multi-instance annotations (e.g. ``multifocal pneumonia''
or ``bibasilar consolidations'') which would typically fall out
of scope for referring expressions, but were included at the radiologist's
discretion in approximately 46 of the annotations. Considering the
case of ``bibasilar consolidations'', the ground truth annotation
indicates two symmetrical boxes on each lung lobe. Such annotations
are especially challenging for the one-stage method as it does not
consider the multi-instance problem. 

The ablation study on different levels of NL information is given
in Tab. \ref{tab3}. Note, the table retains the same architecture
and training procedure from the LITERATI description in sec. \ref{subsec:Network-architecture}.
\begin{table}
\begin{centering}
\caption{Ablation with differing NL information supplied during training}
\label{tab3} %
\begin{tabular}{|c|c|c|c|c|c|}
\hline 
IOU & 0.1 & 0.2 & 0.3 & 0.4 & 0.5\tabularnewline
\hline 
\hline 
Scene label & 0.337 & 0.123 & 0.048 & 0.012 & 0.000\tabularnewline
\hline 
Referring expression & \textbf{0.349} & \textbf{0.125} & 0.051 & 0.014 & 0.002\tabularnewline
\hline 
Referring disease emphasis & \textbf{0.349} & \textbf{0.125} & \textbf{0.060} & \textbf{0.024} & \textbf{0.007}\tabularnewline
\hline 
\end{tabular}
\par\end{centering}
\centering{}Weakly supervised experiments with varying language input,
i.e. scene label (``pneumonia''), referring expression (``patchy
right infrahilar opacity''), or referring expression with scene label
disease (``large left pneumonia''). 
\end{table}
 The improvements using detail ranging from the scene level label
to a full disease related sentence show performance gain at high IOUs.
Although the training NL phrasing differs in the ablation, at test
time the phrasing is the same across methods. Since a pretrained fixed
BERT encoder is used, the ablation probes the adaptability from the
NL portion of the architecture to the task. Since the pretrained encoder
is trained on a large generic corpora, it likely retains the NLP pipeline
(named entity recognition, coreference disambiguation, etc.) necessary
for the visual grounding task. In limited experiments (not shown),
fine-tuning on corpora appears to depend granularly on matching the
domain specific corpora with the task.

\section{Conclusion}

We present a weakly supervised vision language method and associated
clinical referring expressions dataset on pneumonia and pneumothorax
chest x-ray images at scale. The clinical reports generate expressions
that isolate discriminative objects inside the images. As parsing
into referring expressions is accurate and mostly independently of
vocabulary (i.e. it's tractable to identify a direct object without
knowing exactly the meaning of the object) \cite{referit}, the referring
phrases represent a valuable source of information during the learning
process. 

Though not necessarily motivated by learning processes in nature,
algorithms including NL bend towards explainable mechanisms, i.e.
the localized image and language pairs form clear concepts. The explainable
nature of visually grounded referring expressions in a clinical setting,
while cogent here, merits further investigation on the lines of workflow
performance. For pure NLP tasks, training on a data distribution closely
matching the testing distribution has encountered success. An appropriately
matching referring expressions dataset may draw from an ontology \cite{chestx8,deeplesion}
or from didactic literature.

The study suggests that vision language approaches may be valuable
for accessing information within clinical reports.\pagebreak{}

\bibliographystyle{splncs04}
\bibliography{literati}

\begin{thebibliography}{10}
\providecommand{\url}[1]{\texttt{#1}}
\providecommand{\urlprefix}{URL }
\providecommand{\doi}[1]{https://doi.org/#1}

\bibitem{DBLP:conf/nips/BergstraBBK11}
Bergstra, J., Bardenet, R., Bengio, Y., K{\'{e}}gl, B.: Algorithms for
  hyper-parameter optimization. In: Shawe{-}Taylor, J., Zemel, R.S., Bartlett,
  P.L., Pereira, F.C.N., Weinberger, K.Q. (eds.) Advances in Neural Information
  Processing Systems 24: 25th Annual Conference on Neural Information
  Processing Systems 2011. Proceedings of a meeting held 12-14 December 2011,
  Granada, Spain. pp. 2546--2554 (2011),
  \url{http://papers.nips.cc/paper/4443-algorithms-for-hyper-parameter-optimization}

\bibitem{mscocoanno}
Brooks, J.: Coco annotator. \url{https://github.com/jsbroks/coco-annotator/}
  (2019)

\bibitem{bertart}
Devlin, J., Chang, M., Lee, K., Toutanova, K.: {BERT:} pre-training of deep
  bidirectional transformers for language understanding. CoRR
  \textbf{abs/1810.04805} (2018), \url{http://arxiv.org/abs/1810.04805}

\bibitem{frcnn}
Girshick, R.B.: Fast {R-CNN}. In: 2015 {IEEE} International Conference on
  Computer Vision, {ICCV} 2015, Santiago, Chile, December 7-13, 2015. pp.
  1440--1448. {IEEE} Computer Society (2015). \doi{10.1109/ICCV.2015.169},
  \url{https://doi.org/10.1109/ICCV.2015.169}

\bibitem{speedacctrade}
Huang, J., Rathod, V., Sun, C., Zhu, M., Korattikara, A., Fathi, A., Fischer,
  I., Wojna, Z., Song, Y., Guadarrama, S., Murphy, K.: Speed/accuracy
  trade-offs for modern convolutional object detectors. CoRR
  \textbf{abs/1611.10012} (2016), \url{http://arxiv.org/abs/1611.10012}

\bibitem{chexpert}
Irvin, J., Rajpurkar, P., Ko, M., Yu, Y., Ciurea{-}Ilcus, S., Chute, C.,
  Marklund, H., Haghgoo, B., Ball, R.L., Shpanskaya, K.S., Seekins, J., Mong,
  D.A., Halabi, S.S., Sandberg, J.K., Jones, R., Larson, D.B., Langlotz, C.P.,
  Patel, B.N., Lungren, M.P., Ng, A.Y.: Chexpert: {A} large chest radiograph
  dataset with uncertainty labels and expert comparison. CoRR
  \textbf{abs/1901.07031} (2019), \url{http://arxiv.org/abs/1901.07031}

\bibitem{mimic-cxr}
Johnson, A.E.W., Pollard, T.J., Berkowitz, S.J., Greenbaum, N.R., Lungren,
  M.P., Deng, C., Mark, R.G., Horng, S.: {MIMIC-CXR:} {A} large publicly
  available database of labeled chest radiographs. CoRR
  \textbf{abs/1901.07042} (2019), \url{http://arxiv.org/abs/1901.07042}

\bibitem{cvprdense}
Johnson, J., Karpathy, A., Fei-Fei, L.: Densecap: Fully convolutional
  localization networks for dense captioning. In: Proceedings of the IEEE
  Conference on Computer Vision and Pattern Recognition (CVPR) (June 2016)

\bibitem{thtang}
Kao, H.: Gradcam on chexnet (Mar 2020),
  \url{https://github.com/thtang/CheXNet-with-localization}

\bibitem{referit}
Kazemzadeh, S., Ordonez, V., Matten, M., Berg, T.L.: Referitgame: Referring to
  objects in photographs of natural scenes. In: Moschitti, A., Pang, B.,
  Daelemans, W. (eds.) Proceedings of the 2014 Conference on Empirical Methods
  in Natural Language Processing, {EMNLP} 2014, October 25-29, 2014, Doha,
  Qatar, {A} meeting of SIGDAT, a Special Interest Group of the {ACL}. pp.
  787--798. {ACL} (2014). \doi{10.3115/v1/d14-1086},
  \url{https://doi.org/10.3115/v1/d14-1086}

\bibitem{thoracicart}
Li, Z., Wang, C., Han, M., Xue, Y., Wei, W., Li, L., Li, F.: Thoracic disease
  identification and localization with limited supervision. CoRR
  \textbf{abs/1711.06373} (2017), \url{http://arxiv.org/abs/1711.06373}

\bibitem{networkinnet}
Lin, M., Chen, Q., Yan, S.: Network in network. In: Bengio, Y., LeCun, Y.
  (eds.) 2nd International Conference on Learning Representations, {ICLR} 2014,
  Banff, AB, Canada, April 14-16, 2014, Conference Track Proceedings (2014),
  \url{http://arxiv.org/abs/1312.4400}

\bibitem{nltk}
Loper, E., Bird, S.: {NLTK:} the natural language toolkit. CoRR
  \textbf{cs.CL/0205028} (2002), \url{https://arxiv.org/abs/cs/0205028}

\bibitem{alarm}
Lyubinets, V., Boiko, T., Nicholas, D.: Automated labeling of bugs and tickets
  using attention-based mechanisms in recurrent neural networks. CoRR
  \textbf{abs/1807.02892} (2018), \url{http://arxiv.org/abs/1807.02892}

\bibitem{corenlp}
Manning, C.D., Surdeanu, M., Bauer, J., Finkel, J.R., Bethard, S., McClosky,
  D.: The stanford corenlp natural language processing toolkit. In: Proceedings
  of the 52nd Annual Meeting of the Association for Computational Linguistics,
  {ACL} 2014, June 22-27, 2014, Baltimore, MD, USA, System Demonstrations. pp.
  55--60. The Association for Computer Linguistics (2014).
  \doi{10.3115/v1/p14-5010}, \url{https://doi.org/10.3115/v1/p14-5010}

\bibitem{ibm-morardi}
Moradi, M., Madani, A., Gur, Y., Guo, Y., Syeda-Mahmood, T.: Bimodal network
  architectures for automatic generation of image annotation from text. In:
  Frangi, A.F., Schnabel, J.A., Davatzikos, C., Alberola-L{\'o}pez, C.,
  Fichtinger, G. (eds.) Medical Image Computing and Computer Assisted
  Intervention -- MICCAI 2018. pp. 449--456. Springer International Publishing,
  Cham (2018)

\bibitem{pytorchart}
Paszke, A., Gross, S., Massa, F., Lerer, A., Bradbury, J., Chanan, G., Killeen,
  T., Lin, Z., Gimelshein, N., Antiga, L., Desmaison, A., K{\"{o}}pf, A., Yang,
  E., DeVito, Z., Raison, M., Tejani, A., Chilamkurthy, S., Steiner, B., Fang,
  L., Bai, J., Chintala, S.: Pytorch: An imperative style, high-performance
  deep learning library. CoRR  \textbf{abs/1912.01703} (2019),
  \url{http://arxiv.org/abs/1912.01703}

\bibitem{chexnet}
Rajpurkar, P., Irvin, J., Zhu, K., Yang, B., Mehta, H., Duan, T., Ding, D.Y.,
  Bagul, A., Langlotz, C., Shpanskaya, K.S., Lungren, M.P., Ng, A.Y.: Chexnet:
  Radiologist-level pneumonia detection on chest x-rays with deep learning.
  CoRR  \textbf{abs/1711.05225} (2017), \url{http://arxiv.org/abs/1711.05225}

\bibitem{squad2}
Rajpurkar, P., Jia, R., Liang, P.: Know what you don't know: Unanswerable
  questions for squad. CoRR  \textbf{abs/1806.03822} (2018),
  \url{http://arxiv.org/abs/1806.03822}

\bibitem{yolo}
Redmon, J., Divvala, S.K., Girshick, R.B., Farhadi, A.: You only look once:
  Unified, real-time object detection. CoRR  \textbf{abs/1506.02640} (2015),
  \url{http://arxiv.org/abs/1506.02640}

\bibitem{yolov3}
Redmon, J., Farhadi, A.: Yolov3: An incremental improvement. CoRR
  \textbf{abs/1804.02767} (2018), \url{http://arxiv.org/abs/1804.02767}

\bibitem{gradcam}
Selvaraju, R.R., Das, A., Vedantam, R., Cogswell, M., Parikh, D., Batra, D.:
  Grad-cam: Why did you say that? visual explanations from deep networks via
  gradient-based localization. CoRR  \textbf{abs/1610.02391} (2016),
  \url{http://arxiv.org/abs/1610.02391}

\bibitem{bertnlppipe}
Tenney, I., Das, D., Pavlick, E.: {BERT} rediscovers the classical {NLP}
  pipeline. CoRR  \textbf{abs/1905.05950} (2019),
  \url{http://arxiv.org/abs/1905.05950}

\bibitem{attentionall}
Vaswani, A., Shazeer, N., Parmar, N., Uszkoreit, J., Jones, L., Gomez, A.N.,
  Kaiser, L., Polosukhin, I.: Attention is all you need. CoRR
  \textbf{abs/1706.03762} (2017), \url{http://arxiv.org/abs/1706.03762}

\bibitem{chestx8}
Wang, X., Peng, Y., Lu, L., Lu, Z., Bagheri, M., Summers, R.M.: Chestx-ray8:
  Hospital-scale chest x-ray database and benchmarks on weakly-supervised
  classification and localization of common thorax diseases. CoRR
  \textbf{abs/1705.02315} (2017), \url{http://arxiv.org/abs/1705.02315}

\bibitem{tienet}
Wang, X., Peng, Y., Lu, L., Lu, Z., Summers, R.M.: Tienet: Text-image embedding
  network for common thorax disease classification and reporting in chest
  x-rays. CoRR  \textbf{abs/1801.04334} (2018),
  \url{http://arxiv.org/abs/1801.04334}

\bibitem{deeplesion}
Yan, K., Wang, X., Lu, L., Summers, R.M.: Deeplesion: Automated deep mining,
  categorization and detection of significant radiology image findings using
  large-scale clinical lesion annotations. CoRR  \textbf{abs/1710.01766}
  (2017), \url{http://arxiv.org/abs/1710.01766}

\bibitem{onestage}
Yang, Z., Gong, B., Wang, L., Huang, W., Yu, D., Luo, J.: A fast and accurate
  one-stage approach to visual grounding. In: 2019 {IEEE/CVF} International
  Conference on Computer Vision, {ICCV} 2019, Seoul, Korea (South), October 27
  - November 2, 2019. pp. 4682--4692. {IEEE} (2019).
  \doi{10.1109/ICCV.2019.00478}, \url{https://doi.org/10.1109/ICCV.2019.00478}

\bibitem{onestageart}
Yang, Z., Gong, B., Wang, L., Huang, W., Yu, D., Luo, J.: A fast and accurate
  one-stage approach to visual grounding. CoRR  \textbf{abs/1908.06354} (2019),
  \url{http://arxiv.org/abs/1908.06354}

\bibitem{cam}
Zhou, B., Khosla, A., Lapedriza, {\`{A}}., Oliva, A., Torralba, A.: Learning
  deep features for discriminative localization. CoRR  \textbf{abs/1512.04150}
  (2015), \url{http://arxiv.org/abs/1512.04150}

\bibitem{wentao}
Zhu, W., Vang, Y.S., Huang, Y., Xie, X.: Deepem: Deep 3d convnets with em for
  weakly supervised pulmonary nodule detection. In: Frangi, A.F., Schnabel,
  J.A., Davatzikos, C., Alberola-L{\'o}pez, C., Fichtinger, G. (eds.) Medical
  Image Computing and Computer Assisted Intervention -- MICCAI 2018. pp.
  812--820. Springer International Publishing, Cham (2018)

\end{thebibliography}

\end{document}